\documentclass[11pt,a4paper]{article}
\usepackage[hyperref]{acl2021}
\usepackage{times}
\usepackage{latexsym}

\usepackage{lingstyle}
\usepackage{avm}

\usepackage{graphicx,epsfig,color}
\usepackage{epstopdf}
\usepackage[latin1]{inputenc}
\usepackage{natbib}
\usepackage{float}
\usepackage{amssymb}
\usepackage{color}
\usepackage{graphicx,epsfig,color}
\usepackage{epstopdf}
\usepackage{longtable}
\definecolor{darkblue}{rgb}{0,0,0.8}
\definecolor{darkgreen}{rgb}{0,0.5,0}
\definecolor{darkred}{rgb}{0.8,0,0}
\definecolor{brown}{rgb}{0.5,0.3,0}

{\obeyspaces\gdef {\ }}
\global\newbox\codebox
\global\newbox\savedcodebox
\gdef\sverbatim{\bgroup\def\endsverbatim{\egroup\egroup\egroup\mbox{\box\codebo\
x}}\def\savecode{\egroup\egroup\egroup\global\setbox\savedcodebox\copy\codebox}\
\def\par{\egroup\vspace{-0.3em}\hbox\bgroup}\tt\obeylines\obeyspaces\global\set\
box\codebox\vbox\bgroup\hbox\bgroup}

\newenvironment{enumerate*}%
  {\begin{enumerate}%
    \setlength{\itemsep}{0pt}%
    \setlength{\parskip}{0pt}}%
  {\end{enumerate}}

\newenvironment{itemize*}%
  {\begin{itemize}%
    \setlength{\itemsep}{0pt}%
    \setlength{\parskip}{0pt}}%
  {\end{itemize}}

\usepackage{microtype}

\aclfinalcopy 

\title{An Abstract Specification of VoxML as an Annotation Language}

\author{Kiyong Lee \\
  Dept. of Linguistics\\
  Korea University, Seoul \\
  {\small \texttt{ikiyong@gmail.com}}\\  \And
  Nikhil Krishnaswamy \\
  Dept. of Computer Science\\
  Colorado State University\\
  {\small \texttt{nkrishna@colostate.edu}} \\  \And
  James Pustejovsky\\
  Dept. of Computer Science\\
  Brandeis University \\
  {\small \texttt{jamesp@brandeis.edu} }
  }

\date{}

\begin{document}
\maketitle
\begin{abstract} 
VoxML is a modeling language used to map natural language expressions into real-time visualizations using commonsense semantic knowledge of objects and events. Its utility has been demonstrated in embodied simulation environments and in  agent-object interactions in situated multimodal human-agent collaboration and communication. It introduces  the notion of object affordance (both  Gibsonian and Telic) from HRI and robotics, as well as the  concept of habitat (an object's context of use) for  interactions between a rational agent  and an object. This paper aims to specify VoxML as an annotation language in general abstract terms. It then shows how it works on annotating linguistic data that express visually perceptible human-object interactions. The annotation structures thus generated will be interpreted against the enriched minimal model created by VoxML as a modeling language while supporting the modeling purposes of VoxML linguistically. 
\end{abstract}

\section{Introduction}

As introduced by \citet{PustejovskyKrishnaswamy:2016},
VoxML is a modeling language encoding the spatial and visual components of an object's conceptual structure.\footnote{VoxML represents a {\it visual object concept structure (vocs)}  modeling language.} It allows for 3D visual interpretations and simulations of objects, motions, and actions as minimal models from verbal descriptions. The  data structure associated with this is called a {\it voxeme}, and the library of voxemes is referred to as a {\it voxicon}.


VoxML elements are conceptually grounded by  a conventional inventory of semantic types \citep{pustejovsky1995,pustejovsky2019lexicon}. They are also enriched  with a representation of how and when an object affords interaction with another object or an agent. This is a natural  extension of Gibson's notion of object affordance \cite{Gibson:1977} to functional and goal-directed aspects of Generative Lexicon's Qualia Structure \cite{Pustejovsky:2013,pustejovsky2021embodied}, 
and is situationally grounded within a semantically interpreted 4D  simulation environment (temporally interpreted 3D space), called VoxWorld \cite{mcneely2019user,krishnaswamy2022voxworld}.

VoxML has also been proposed for annotating visual information as part of the ISO 24617 series of international standards on semantic annotation schemes, such as ISO-TimeML \citep{ISO:2012} and ISO-Space \citep{ISO:2020}. VoxML, as an annotation language, should be specified in abstract terms, general enough to be interoperable with other annotating languages, especially as part of such ISO standards, while licensing various implementations in concrete terms. In order to address these requirements, this paper aims to formulate an abstract syntax of VoxML based on a metamodel.  It develops as follows: Section \ref{motivate}, Motivating VoxML as an Annotation Language, Section \ref{spec}, Specification of an Annotation Scheme, based on VoxML, Section \ref{interpret}, Interpretation of Annotation-based Logical Forms with respect to the VoxML Minimal Model, and Section \ref{conclude}, Concluding Remarks.

\section{\label{motivate} Motivating VoxML as an Annotation Language}

Interpreting actions and motions requires situated background information about their agents or related objects, occurrence conditions, and enriched lexical information. The interpretation of base annotation structures, anchored to lexical markables for annotating visual perceptions, depends on various sorts of parametric information besides their associated dictionary definitions.

A significant part of any model for situated communication is an encoding of the semantic type, functions, purposes, and uses introduced by the ``objects under discussion''. For example, a semantic model of 
perceived {\it  object teleology}, as introduced by Generative Lexicon (GL) with the Qualia Structure, for example, \cite{pustejovsky1995}, as well as {\it  object affordances} \cite{Gibson:1977} is useful to help ground expression meaning to speaker intent. As an illustration, consider first how such information is encoded and then exploited in reasoning.  Knowledge of objects can be  partially contextualized  through their {\it qualia structure} \cite{pustejovsky1993lexical}, where each Qualia role can be seen as  answering  a specific question about the object it is bound to:  {\it Formal}, the {\sc is-a} relation;
{\it Constitutive}, an object {\sc part-of} or  {\sc made-of} relation; {\it Agentive}, the object's {\sc created-by} relation; and 
{\it Telic}: encoding information on purpose and function (the {\tt used-for} or {\sc functions-as} relation).

While such information is needed for compositional semantic operations and inferences in conventional models, it falls short of providing a  representation for the {\it situated grounding} of events and their participants or of any expressions between individuals involved in a communicative exchange.
VoxML provides just such a representation. It further encodes objects with rich semantic typing and action affordances and actions themselves as multimodal programs, enabling contextually salient inferences and decisions in the environment.
To illustrate this, consider the short narrative in (\ref{ex1}) below.

\enumsentence{ 
Mary picked up the glass from the table and put it in the dishwasher 
to wash and dry it.
\label{ex1}}

\noindent VoxML provides the means to better interpret these events as situationally grounded in  interactions between an agent and objects in the world.  

In order to create situated interpretations for each of these events, there must be some semantic encoding associated with how the objects relate to each other physically and how they are configured to each other spatially. For example, if we associate the semantic type of ``container'' with glass, it is situationally important to know how and when the container capability is activated: i.e., the orientation information is critical for enabling the use or function of the glass {\it qua} container. VoxML encodes these notions that are critical for Human-Object Interaction as:  {\it what} the function  associated with an object is (its affordance), and just as critically,  {\it when} the affordance is active (its habitat). It also explicitly encodes the dynamics of the events bringing about any object state changes in the environment, e.g., change in location, time, and attribute. 

\section{\label{spec} Specification of the Annotation Scheme}

\subsection{Overview}
\begin{figure}[ht]
\centering
\includegraphics[scale=0.75]{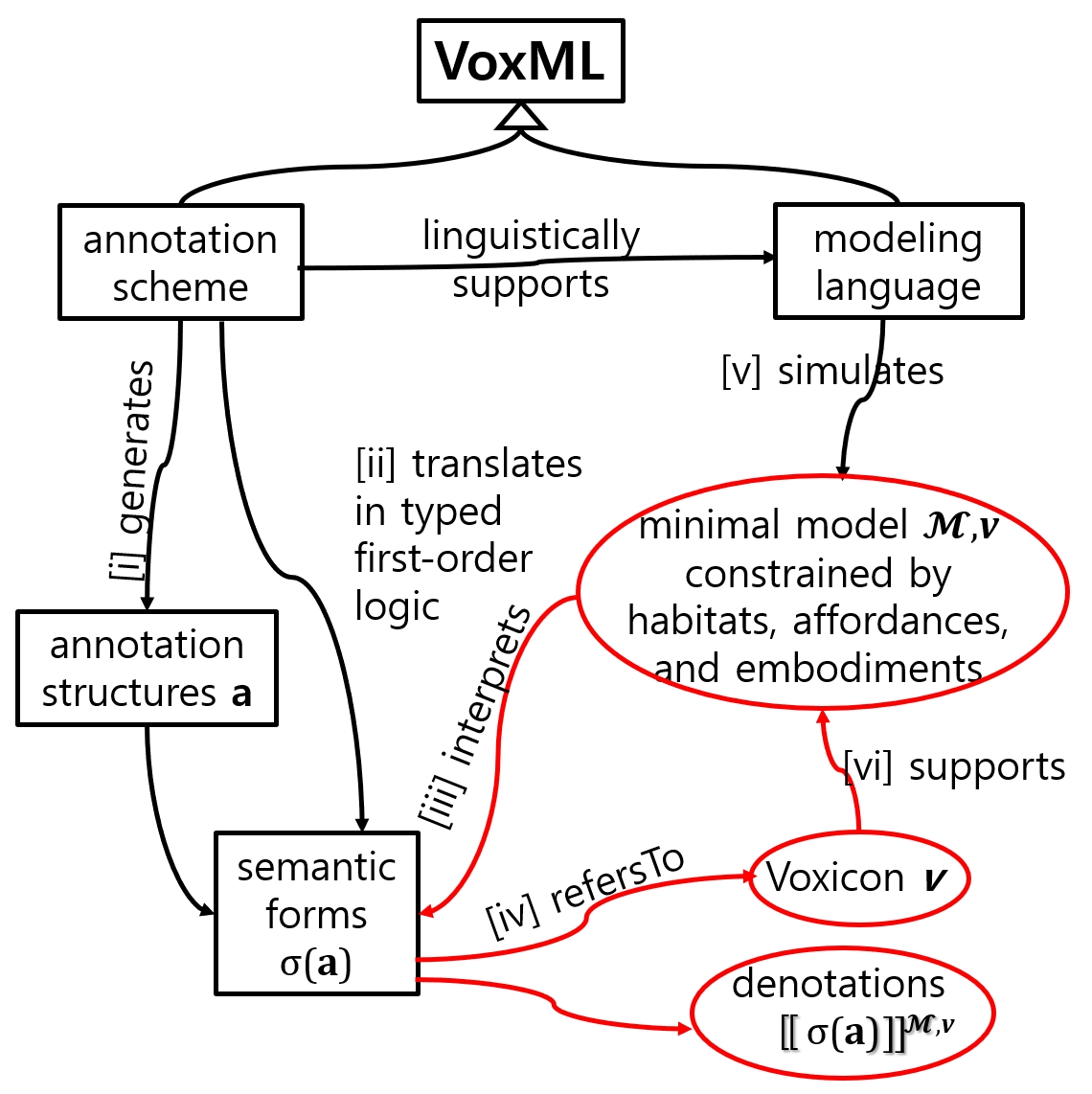}
\caption{How VoxML operates}
\label{twoUses}
\end{figure}

VoxML is primarily a modeling language for simulating actions in the visual world. Still, it can also be used as a markup language for (i) annotating linguistic expressions involving human-object interactions, (ii) translating annotation structures in shallow semantic forms in typed first-order logic, and then (iii) interpreting with the minimal model simulated by VoxML by referring to the voxicon, or set of voxemes, as shown in Figure \ref{twoUses}. 

This section formally specifies the VoxML-based annotation scheme, with a metamodel (\ref{meta}), an abstract syntax (\ref{ASyn}), a concrete representation of annotation structures (\ref{represent}), and their translation to semantic forms in typed first-order logic (\ref{annSem}).

\subsection{\label{meta} Metamodel of the VoxML-based Annotation Scheme}

\begin{figure}[ht]
\centering
\includegraphics[scale=1.0]{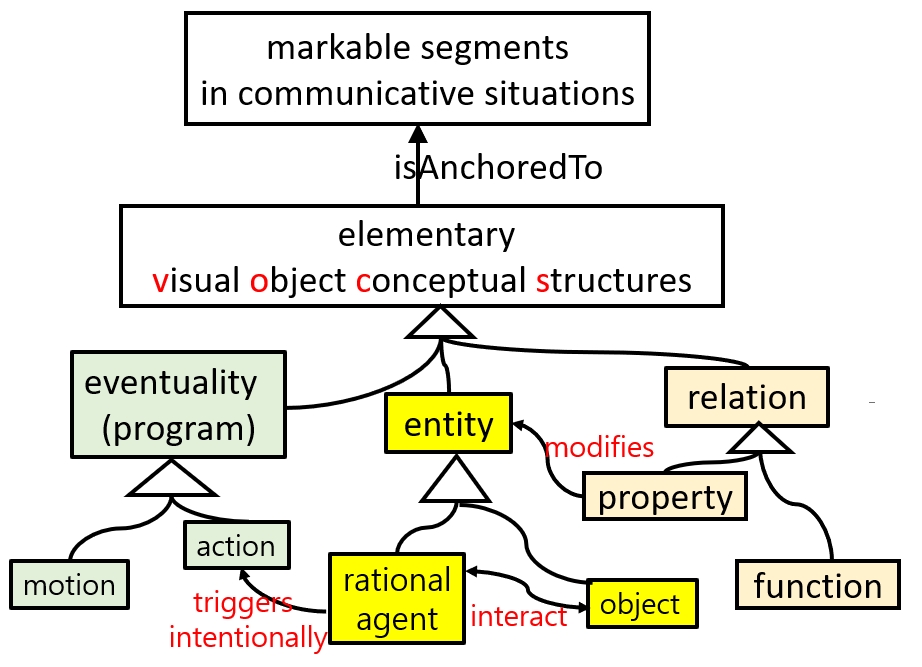}
\caption{Metamodel of VoxML}
\label{metamodel}
\end{figure}

A metamodel graphically depicts the general structure of a markup language. 
As pointed out by \citet{Bunt:2022}, a metamodel makes the specification of annotation schemes
intuitively more transparent, thus becoming a {\it de facto} requirement for constructing semantic annotation schemes. The metamodel, represented by Figure \ref{metamodel}, focuses on interactions between entities (objects) and humans, while the {\it dynamic paths}, triggered by their actions, trace the visually perceptible courses of those actions. The VoxML-based annotation scheme, thus represented, is construed to annotate linguistic expressions for human-object interactions (cf. \citet{HenleinEtAl:2023}). 

We view the VoxML model or world as inhabited by only three categories of entities: {\it event (program):action}, {\it object}, and {\it relation}. Each of them has subcategories, as represented by the hollow triangles in Figure \ref{metamodel}.\footnote{In UML, a hollow triangle represents a subcategorization relation.} Because of its key role in VoxML, category {\it action} is introduced as a subcategory of category {\it event}. This model represents a {\it small minimal} world, focused on actions, (physical) objects, and their interrelations,
which together constitute the larger ontology such as SUMO \citep{NilesPease:2001}. Unlike other types of eventuality, agents intentionally trigger actions, and these agents can be humans or other rational agents. These agents also interact with objects as participants in actions. 

Category {\it relation} has two subcategories, {\it property} and {\it function}. As unary relations, properties modify entities (objects), as in {\it big table}. Functions are particular relations mapping one object to another. The function $loc$ for {\it localization}, for instance, maps physical objects (e.g., {\it table}) to spatial locations where some other objects like apples can be placed.
As introduced by \citet{Katz:2007}, the runtime function $\tau$ maps eventualities to times such that $\tau(e)$ refers to the occurrence time of the event $e$. We may also introduce a function $seq$ that forms paths by ordering pairs $t@l$ of a time $t$ and a location $l$. The VoxML annotation language has no category such as location, time, or path, but can introduce time points to discuss, for instance, their temporal ordering: e.g., $\tau(e_1) \prec \tau(e2)$. Binary or any other $n$-ary relations, such as {\it in} or {\it between}, are of category {\it relation} and are also introduced into VoxML.  

VoxML, as a modeling language, views physical objects and actions as forming visually perceptible conceptual structures called {\it voxemes}. Applied to language and its constituent expressions, the VoxML-based annotation scheme takes them as {\it markables}, anchored to a word, an image, a gesture, or anything from communicative actions that consist of verbal descriptions, gestures, and surrounding backgrounds. 

\subsection{\label{ASyn} Abstract Syntax}

An abstract syntax defines a specification language and rigorously formulates its structures. 
In constructing natural language grammars \citep{Lee:2016, Lee:2023}, the abstract syntax of a semantic annotation scheme is defined as a tuple in set-theoretic terms. The abstract syntax $\mathcal{AS}yn_{voxml}$ of the VoxML-based annotation scheme is also defined as a set-theoretic tuple, as in Definition \ref{ASynD}:

{\small
\enumsentence{\label{ASynD} Definition of $\mathcal{AS}yn_{voxml}$:\\
Given a finite set $D$, or data, of communicative segments in natural language, the abstract syntax $\mathcal{AS}yn_{voxml}$ of VoxML is defined to be a triplet {\small $<$$M, C, @$$>$}, where:
\begin{itemize}
\item $M$ is a nonnull subset of $D$ that contains (possibly null or non-contiguous) strings of communicative segments, called {\it markables}, each delimited by the set $B$ of  base categories.
\item $C$ consists of base categories $B$ and relational categories $R$: 
\begin{itemize}
\item Base categories $B$ and their subcategories, as depicted in Figure \ref{metamodel}: [i] {\it event}:{\it action}, [ii] {\it entity (object)} and [iii] {\it relation}:\{{\it property}, {\it function}\}.
\item Relational categories $R$: unspecified for $\mathcal{AS}yn_{voxml}$.
\end{itemize}
\item $@_{cat}$ is a set of assignments from attributes to values specified for each category $cat$ in $C$.
\end{itemize}
}
}

\noindent 
For every base category $cat$ in $B$, the assignment $@_{cat}$ has the following list of attributes as required to be assigned a value:

{\small
\enumsentence{\label{assignAction} {\bf Assignment} $@_{cat}$ in Extended BNF:\\
{\small\tt
attributes = \\
identifier, target, type, pred;\\
identifier = categorized prefix\\
\hspace*{0.5em} + a natural number;\\
target = markable;\\ 
type = CDATA; \\
pred = CDATA|null; \\
\hspace*{0.5em} (* predicative content *)
}
}}

\noindent
Each category may have additional required or optional attributes to be assigned a value. For instance, the assignment $@_{action}$ is either a {\it process} or {\it transition} type. Category {\it action} has the attribute {\small\tt @agent}, which triggers it. 

\subsection{\label{represent} Representing Annotation Structures}
The annotation scheme, such as $\mathcal{AS}_{voxml}$, generates annotation structures based on its abstract syntax. These annotation structures have two substructures: {\it anchoring} and {\it content} structures. In pFormat\footnote{pFormat is a predicate-logic-like annotation format
for replacing XML, thus being constrained to introduce embedded structures into annotations.}, 
these two structures are represented differently by representing anchoring structures by their values only,
but content structures as attribute-value pairs.

The first part of Example (\ref{ex1}) is annotated as follows:

\eenumsentence{\item Base-segmented Data:\\
{\small Mary$_{x1,w1}$ picked up$_{e1,w2-3}$ the glass$_{x2,w5}$ from$_{r1,w6}$ the table$_{x3,w8}$.}
\item Annotation Structures:\\
{\small\tt {\bf object}(x1, w1, \\
type="human", pred="mary")\\ 
{\bf action}(e1, w2-3, \\
type="transition", pred="pickUp",  \\
agent="\#x1", physObj="\#x2")\\
{\bf object}(x2, w5,\\ 
type="physobj", pred="glass")\\ 
{\bf relation}(r1, w6,\\
type="spatial", source="\#x3")\\
{\bf object}(x3, w8,\\
type="physobj", pred="table")
}
}\label{anno-1}
\noindent
In base-segmented data, each markable is identified by its anchoring structure $<$$cat_i,w_j$$>$ (e.g., {\tt x1, w1}), where $cat_i$ is a categorized identifier 
and $w_j$ is a word identifier. The agent which triggered the action of picking up the glass is marked as Mary$_{x1}$, and the object glass$_{x2}$ is related to it.

\paragraph{Interoperability} is one of the adequacy requirements for an annotation scheme. Here, we show how the VoxML-based annotation scheme is interoperable with other annotation schemes, such as ISO-TimeML \citep{ISO:2012} and the annotation scheme on anaphoric relations (see \citet{Lee:2017} and \citet{ISO:2019}). The rest of Example (\ref{ex1}) can also be annotated with these annotation schemes. It is first word-segmented, while each markable is tagged with a categorized identifier and a word identifier as in (\ref{bSeg-2}):

\eenumsentence{\item Primary Data:\\
Mary picked up the glass from the table and put it in the dishwasher to wash and dry it.
\item Base-segmented Data:\\
{\small Mary$_{x1,w1}$ [picked up]$_{e1,w2-3}$ the\\
glass$_{x2,w5}$ from$_{r1,w6}$ the table$_{x3,w8}$ and\\
put$_{e2,w10}$ it$_{x4,w11}$ 
in$_{r2,w12}$ the dishwasher$_{x5,w14}$ \\
to wash$_{e3,w16}$ and dry$_{e4,w18}$ it$_{x6,w19}$.
}
}\label{bSeg-2}

Second, each markable is annotated as in (\ref{anno-2}):

\enumsentence{Elementary Annotation Structures:\\
{\small\tt {\bf action}(e2, w10\\
type="transition", pred="put"\\
agent="\#x1", relatedTo="\#x4")\\
{\bf object}(x4, w11,\\
type="unknown", pred="pro")\\
{\bf relation}(r2, w12\\
type="spatial", pred="in")\\
{\bf object}(x5, w14, \\
type="physobj, artifact",\\
pred="dishwasher")\\
{\bf action}(e3, w16,\\
type="process", pred="wash",\\
agent="\#5, theme="\#x6")\\
{\bf action}(e4, w18,\\
type="process", pred="dry",\\
agent="\#x5", theme="\#x6")\\
{\bf object}(x6, w19,\\
type="unknown", pred="pro")
}
}\label{anno-2}
\noindent
The first two actions {\it pick up} and  {\it put} are triggered by the human agent {\it Mary}, whereas the actions of {\it wash} and {\it dry} are triggered by the dishwasher, which is not human. 

The annotation scheme $\mathcal{AS}_{voxM\hspace*{-0.15em}L}$ for actions annotates the temporal ordering of these four actions by referring to ISO-TimeML, as in (\ref{tLink}):

\eenumsentence{\item Temporal Links ({\tt tLink}):\\
{\small\tt {\bf tLink}(tL1, eventID="\#e2", relatedToEventID="\#e1", relType="after")}\\
{\small\tt {\bf tLink}(tL2, eventID="\#e3", relatedToEventID="\#e2", relType="after")}\\
{\small\tt {\bf tLink}(tL3, eventID="\#e4", relatedToEventID="\#e3", relType="after")}
\item Semantic Representation:\\
{\small [$pickU$\hspace*{-0.2em}$p(e_1), put(e_2), wash(e_3), dry(e_4),\\
\tau(e_1)\prec \tau(e_2) \prec \tau(e_3) \prec \tau(e_4)$]}\footnote{These semantic forms can be represented in DRS validly. See \citet{Lee:2023}.}
}\label{tLink}

The annotation scheme $\mathcal{AS}_{voxM\hspace*{-0.15em}L}$ can also refer to the  subordination link ({\tt sLink}) in ISO-TimeML \citep{ISO:2012} to annotate subordinate clauses such as {\it to wash and dry it} in Example (\ref{ex1}). 

\eenumsentence{\item Subordination Link ({\tt sLink}):\\
{\small\tt {\bf sLink}(sL1, eventID="\#e2", \\relatedTo="\{\#e3,\#e4\}", relType="purpose")}
\item Semantic Representation:\\
{\small $[put(e_2), wash(e_3), dry(e_4), \\
purpose(e_2,\{e_3,e_4\})]$}
}\label{sLink}
\noindent
The subordination link (\ref{sLink}) relates the actions of {\it wash} and {\it dry} to the action of {\it put} by annotating that those actions were the {\it purpose} of {\it putting} the glass in the dishwasher.  

By referring to the annotation schemes proposed by \citet{Lee:2017} or \citet{ISO:2019}, the VoxML-based annotation scheme can annotate the anaphoric or referential relations
involving pronouns. The two occurrences of the pronoun {\it it} refer to the
noun {\it the glass} are annotated as in (\ref{anno-3}):

\eenumsentence{\item Annotation of Coreferential Relations:\\
{\small\tt {\bf object}(x2, w5, \\
type="physobj, artifact", pred="glass")\\
{\bf anaLink}(aL1, x4, x2, identity)\\
{\bf anaLink}(aL2, x6, x2, identity)}
\item Semantic Representation:\\
(i) {\small $\sigma(x2) := [glass(x_2)], \\
\hspace*{1.5em} \sigma(aL1) := [x_4$=$x_2], \\
\hspace*{1.5em} \sigma(aL2) := [x_6$=$x_1]$}\\
(ii) {\small $[glass(x_2), x_4$=$x_2, x_6$=$x_2]$}
}\label{anno-3}
\noindent
Semantic Representation (ii) is obtained by unifying all the semantic forms in (i). It says that the two occurrences of the pronoun {\it it} both refer to the glass. 

\subsection{\label{annSem} Annotation-based Semantic Forms}

The annotation scheme translates each annotation structure {\bf a$_{\ref{anno-1}}$}
into a semantic form $\sigma$({\bf a}$_{\ref{anno-1}}$), as in (\ref{semForms}).

\eenumsentence{\item Base Semantic Forms $\sigma$:\footnote{As noted earlier, DRS \citep{KampReyle:1993} represents these semantic forms in an equivalent way.} \\
{\small
$\sigma(x1)$ := $\{x_1\}[human(x_1), mary(x_1)]$\\
$\sigma(x2)$ := $\{x_2\}[physObj(x_2), \\
              \hspace*{6em} glass(x_2)]$\\
$\sigma(x3)$ := $\{x_3\}[physObj(x_3), \\ 
              \hspace*{6em}table(x_3)]$\\
$\sigma(e1)$ := $\{e_1\}[action(e_1),\\
              \hspace*{6em}transition(e_1), \\
              \hspace*{6em} pickU\hspace*{-0.2em}p(e_1), \\ 
              \hspace*{6em} agent(e_1,x_1), \\
              \hspace*{6em} theme(e_1,x_2)]$\\
$\sigma(r1)$ := $\{r_1\}[relation(r_1), \\
              \hspace*{6em} source(r_1,x_3)]$
              }
\item Composition of the Semantic Forms:\\
{\small $\sigma$({\bf a$_4$}) := $\oplus\{\sigma(x1), \sigma(x2), \sigma(x3),
\sigma(e1),\sigma(r1)\}$}
}\label{semForms}
\noindent
By unifying all of the semantic forms in (\ref{semForms}a), we obtain the semantic form $\sigma(\bf a_1)$
of the whole annotation structure ${\bf a}_1$. This semantic form roughly states that Mary picked up a glass (see $\sigma(e1)$), which moved away from the table. This interpretation is too shallow to view how Mary's picking up the glass from the table happened. It was on the table, but now it is no longer there. It is in the hand of Mary, who grabbed it. It didn't move by itself, but its location followed the path of the motion how Mary's hand moved.  

\subsection{Interpreting Annotation-based Semantic Forms}

To see the details of the whole motion, as described by Example (\ref{ex1}a), we must know the exact sense of the verb {\it pick up}. WordNet Search - 3.1 lists 16 senses, most rendered when the verb is used with an Object as a transitive verb.  Picking up a physical object like a glass or a book means taking it up by hand, whereas picking up a child from kindergarten or a hitchhiker on the highway means taking the child home or giving the hitchhiker a ride. Such differences in meaning arise from different agent-object interactions. The VoxML-based annotation scheme refers to Voxicon that consists of voxemes and interprets the annotation-based semantic forms, such as (\ref{semForms}), with respect to a VoxML model.

\section{\label{interpret}Interpretation with respect to the VoxML Minimal Model}
Voxemes in VoxML create a minimal model. Each of the annotation-based semantic forms, as in (\ref{semForms}),
is interpreted with respect to this minimal model by referring to its respective voxemes. 

\subsection{Interpreting Objects}
There are four objects mentioned in Example (\ref{ex1}): $mary(x_1)$, $glass(x_2)$,
$table(x_3)$, and $dishwasher(x_5)$.\footnote{The variables $x_4$ and $x_6$ are assigned to the two occurrences of the pronoun {\it it}.}  The semantics forms in (\ref{semForms}) say very little. For instance, the semantic form $\sigma(x2)$ of the markable {\it glass} in (\ref{semForms}) says it is a {\it physical object} but nothing else. 

In addition to the lexical information, as given by its annotation structure and corresponding semantic form, each entity of category {\it object} in VoxML is enriched with information with the elaboration of [i] its {\it geometrical type}, [ii] the {\it habitat} for actions, [iii] the {\it affordance structures}, both Gibsonian and telic, and [iv] the agent-relative {\it embodiment}. 

\begin{figure}[ht]
\centering
\includegraphics[scale=0.43]{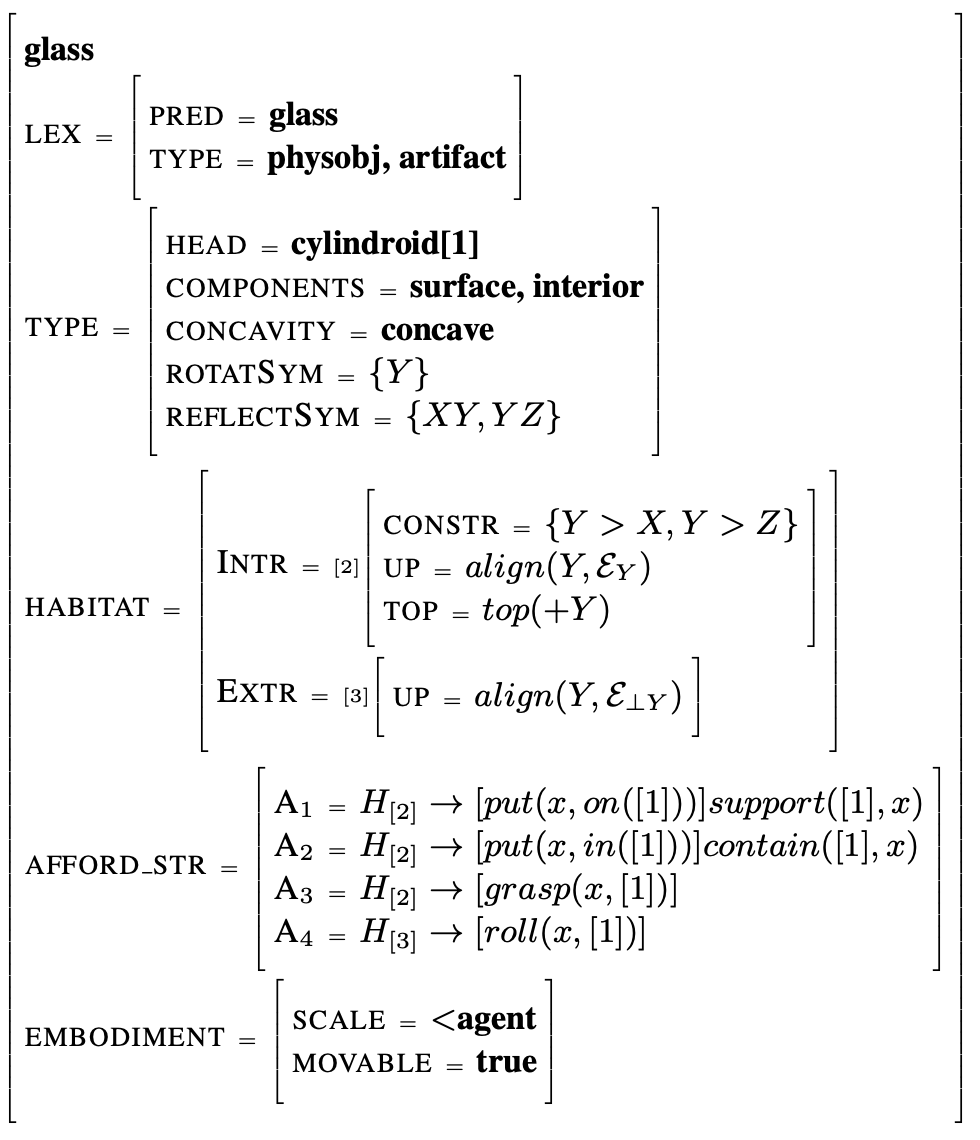}
\caption{VoxML representation for object {\it glass}}
\label{glass}
\end{figure}

In a voxicon, such information is represented in a typed feature structure. An example is given in Figure \ref{glass} for the object {\it glass}.\footnote{ Taken from the Voxicon in  \citet{KrishnaswamyPustejovsky:2020}.}

\noindent
The {\sc type} structure in Figure \ref{glass} contains definitions of rotational symmetry {\sc rotatSym} and reflectional symmetry {\sc reflSym}. The rotational symmetry {\sc rotatSym} of a shape gives the major axis of an object such that when the object is rotated around that axis for some interval of less than or equal to 180 $^{\circ}$, the shape of the object looks the same. Examples of shapes with rotational symmetry are {\it circle}, {\it triangle}, etc.  The reflectional symmetry {\sc reflSym} is a type of symmetry which is with respect to reflections across the plane defined by the axes listed, e.g., a {\it butterfly} assuming vertical orientation would have reflectional symmetry across the YZ-plane.

\begin{figure}[ht]
\centering
\includegraphics[scale=.75]{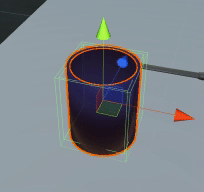}
\caption{Rendering of object {\it glass} (cf. Figure \ref{glass}) showing orthogonal axes.}
\label{glass-render}
\end{figure}

Figure \ref{glass-render} shows a 3D rendering of a glass object as defined by the structure Figure \ref{glass}, taken from the VoxWorld platform \cite{pustejovsky2017object,krishnaswamy2022voxworld}. The object is shown with the 3 major orthogonal axes of the 3D world  The green axis is the Y-axis, which is the axis of rotational symmetry.  The glass is also symmetric across the XY-plane (defined by red and green axes) and the YZ-plane (defined by the green and blue axes).

Under the {\sc habitat} structure in Figure \ref{glass}, the variables $X$, $Y$, and $Z$ correspond to extents in standard Cartesian coordinates, representing the  dimensions, such as areas, required to represent 3D objects in space. From these areas, the radii or circumferences of the bottom and the top areas and the height of the glass are obtainable. Note that the top of a glass has its top area open as a container. Unlike the solid cylindroid, the glass consists of two sheets for the closed bottom and the side such that the circumference of the top area only stands for the width of the side sheet. Note also that the size of the circumference of the top $Y$, which is the brim of a glass, may equal or be larger than that of the bottom $X$. 

The {\it habitat} describes environmental and configurational constraints that are either inherent to the object (``intrinsic'' habitats, such as a glass having an inherent top, regardless of its placement in the environment), or required to execute certain activities with it (``extrinsic'' habitats, such as a glass needing to be placed on its side to be rollable).


This representation  provides the necessary information for its full interpretation. It says the object is glass, a physical artifact having the shape of a concave cylindroid and other geometrical features. 
It should be standing concave upward to hold liquid. Thus, it can be placed on the table, contain water or wine, and be grasped by a hand. It may roll if it falls sideways, but it does only if it does not have something like a handle or is not designed like a wine glass. The embodiment says it is smaller than the one holding it and can move. 

\subsection{Interpreting Agents}

A voxeme for an agent may refer to an actual human agent or an AI agent of any form (humanoid, robotic, or without distinct form). Other entities, or {rational agents}, may function as agents as long as they are capable of executing actions in the world~\cite{krishnaswamy2017monte,pustejovsky2017object}  Examples developed using the VoxWorld platform include collaborative humanoid agents that interact with humans and objects, including interpreting VoxML semantics in real time to exploit and learn about object affordances~\cite{krishnaswamy2017communicating,krishnaswamy2020diana,krishnaswamy2022affordance}, navigating through environments to achieve directed goals~\cite{krajovic2020situated}, and also self-guided exploration where the VoxML semantics ``lurk in the background'' for the agent to discover through exploratory ``play''~\cite{ghaffari2022detecting,ghaffari2023grounding}. The physical definition of agents conditions their actions \cite{pustejovsky2021embodied}. For instance, a humanoid agent with defined {\it hand} $\sqsubseteq_c$ {\it arm} $\sqsubseteq_c$ {\it torso} is enabled to execute the act of grasping, while a robotic agent defined with {\it wheels} $\sqsubseteq_c$ {\it chassis} $\sqsubseteq_c$ {\it self} is enabled for the act of locomotion. This has implications for the semantics of how the agent is interacted with: the humanoid can {\it pick up} objects while the robot can {\it go to} them.

\subsection{Interpreting Actions as Programs}
Actions are viewed as {\it programs} that can formally implement them as processes, (dynamic) sequences of sub-events or states, recursions, algorithms, and execution (see \citet{ManiPustejovsky:2012} and \citet{BergEtAl:2010}).

The voxemes for actions are much simpler than those for objects. They consist of three attributes: [i] Lex for lexical information, [ii] Type for argument structure, and [iii] Body for subevent structure. The information conveyed by [i] and [ii] is provided by the annotation structures for predicates with their attributes @type, @pred, @agent, and @physObj. 

\enumsentence{Annotation Structure:\\
{\small\tt {\bf action}(a1, w2-3, \\
type="transition", pred="pickUP", \\
agent="\#x1", physObj="\#x2")}
}\label{anno-pickUp}

As being of  type {\it transition}, the action of picking up involves two stages
of a motion, [i] the initial stage of {\it grasping} the glass and [ii] the ensuing process of {\it moving} to some direction while {\it holding} it.  This involvement is stated by part of the voxeme for the predicate {\it pick up}, as in (\ref{embodiment}):\footnote{This information is derived from the voxeme for {\it lift} in \citet{KrishnaswamyPustejovsky:2020} and applied to the predicate {\it pick up}.} 

\enumsentence{Embodiment for {\it pick up}:\\
a. E$_1$ = {$grasp(x,y)$}\\
b. E$_2$ = {$[while(hold(x,y)$,\\
        \hspace*{2.1em} $move(x,y,vec(\mathcal{E}_Y)))]$}
}\label{embodiment}
\noindent
The embodiment E$_2$ states that the agent $x$ moves the glass $y$, as her hand and arm move together, along the path or vector $\mathcal{E}_Y$ while holding it (see \citet{HarelEtAl:2000} for {\it while} programs or {\it tail recursion}).

\subsection{Interpreting the Role of Relations}

The preposition {\it from} functions as a spatial relation between the object {\it glass} and the table on which it was located and supported. Then, as the hand of the agent {\it Mary} holding the glass moves, the glass is no longer on the table but moves away along the path that the hand moves. Hence, the relation {\it from} marks the initial point of that path or vector. 

\section{\label{conclude} Concluding Remarks}
The paper specified the VoxML-based annotation scheme in formal terms. The example of the action of Mary picking up a glass from the table showed how that particular example was annotated and how its logical forms were interpreted with a VoxML model
while referring to the voxicon. Each voxeme in the Lexicon, especially that of objects, contains information enriched with the notions of habitat, affordance, and embodiment. As the voxicon develops into a full scale, the task of interpreting annotated language data involving complex interactions between humans and objects
can easily be managed. 

For purposes of exposition, the discussion here focused on the annotation of one short narrative in English involving one verb, {\it pick up}, and one object, {\it glass}. The proposed VoxML-based annotation scheme needs to be applied to large data with a great variety to test the effectiveness of interpreting its annotation structures and corresponding semantic forms against the VoxML model.
At the same time, such an application calls for the need to enlarge the size and variety of the voxicon for modeling purposes as well. The evaluation of the VoxML-based annotation scheme and the extension of the voxicon remain as future tasks.    

\section*{Acknowledgments}
This paper has been revised with the constructive comments from four reviewers. We thank them all for their suggestions.

\bibliographystyle{acl_natbib}
\bibliography{spec2023}

\end{document}